\documentclass[conference]{IEEEtran}

\usepackage{cite}
\usepackage{amsmath,amssymb,amsfonts}
\usepackage{algorithm}
\usepackage{algorithmic}
\usepackage{graphicx}
\usepackage{textcomp}
\usepackage{xcolor}
\usepackage{hyperref}
\usepackage{tabularx}
\usepackage{adjustbox}
\usepackage{multirow}
\usepackage{tabu}
\usepackage{svg}
\usepackage{booktabs}
\usepackage{numprint}

\usepackage{tikz}

\definecolor{tabR}{HTML}{9C0006}
\definecolor{tabG}{HTML}{006100}
\definecolor{tabY}{HTML}{FFD700}

\newcolumntype{Y}{>{\centering\arraybackslash}X}

\def\BibTeX{{\rm B\kern-.05em{\sc i\kern-.025em b}\kern-.08em
    T\kern-.1667em\lower.7ex\hbox{E}\kern-.125emX}}

\begin{document}

\title{Enhancing Once-For-All: A Study on Parallel Blocks, Skip Connections and Early Exits}


\author{

   \IEEEauthorblockN{Simone Sarti}
       \IEEEauthorblockA{
       \textit{DEIB, Politecnico di Milano}\\
       Milan, Italy \\
       simone.sarti@mail.polimi.it
   }
   
   \and
   
   \IEEEauthorblockN{Eugenio Lomurno}
   \IEEEauthorblockA{
       \textit{DEIB, Politecnico di Milano}\\
       Milan, Italy \\
       eugenio.lomurno@polimi.it
   }

   \and
   
   \IEEEauthorblockN{Andrea Falanti}
   \IEEEauthorblockA{
       \textit{DEIB, Politecnico di Milano}\\
       Milan, Italy \\
       andrea.falanti@polimi.it
   }

   \and
   
   \IEEEauthorblockN{Matteo Matteucci}
   \IEEEauthorblockA{
       \textit{DEIB, Politecnico di Milano}\\
       Milan, Italy \\
       matteo.matteucci@polimi.it
   }
}

\maketitle

\begin{abstract}
The use of Neural Architecture Search (NAS) techniques to automate the design of neural networks has become increasingly popular in recent years. 
The proliferation of devices with different hardware characteristics using such neural networks, as well as the need to reduce the power consumption for their search, has led to the realisation of Once-For-All (OFA), an eco-friendly algorithm characterised by the ability to generate easily adaptable models through a single learning process.
In order to improve this paradigm and develop high-performance yet eco-friendly NAS techniques, this paper presents OFAv2, the extension of OFA aimed at improving its performance while maintaining the same ecological advantage.
The algorithm is improved from an architectural point of view by including early exits, parallel blocks and dense skip connections.
The training process is extended by two new phases called Elastic Level and Elastic Height. A new Knowledge Distillation technique is presented to handle multi-output networks, and finally a new strategy for dynamic teacher network selection is proposed.
These modifications allow OFAv2 to improve its accuracy performance on the Tiny ImageNet dataset by up to 12.07\% compared to the original version of OFA, while maintaining the algorithm flexibility and advantages.
\end{abstract}

\begin{IEEEkeywords}
OFAv2, Once-For-All, Early Exits, Parallel Blocks, Dense Skip Connections, Knowledge Distillation, Extended Progressive Shrinking
\end{IEEEkeywords}

\section{Introduction}\label{sec:introduction}

The field of deep learning has recently seen a surge in popularity in the computer science community, as neural networks can exploit large amounts of data to automatically extract relevant features and produce state-of-the-art results.
The evolution of neural networks has come a long way from the classic feed-forward neural networks to convolutional neural networks~(CNNs).
However, until recently, networks were typically designed by artificial intelligence experts, using the knowledge gained from previous model designs to understand which patterns and settings worked best.
The advent of neural architecture search (NAS) techniques has changed this state of affairs. In fact, many of today's best networks have been found through the use of specially designed algorithms that can automatically create and/or extract the best neural network to solve a given task, with little or no human intervention in the process.

However, one of the main drawbacks of conventional NAS techniques is the massive use of computational resources and the limitation in producing generic and reusable architectures.
In particular, models developed using NAS, while powerful, are severely constrained by the task for which they have been optimised, as well as by the nature of the hardware, making it impossible to produce architectures suitable for, for example, mobile and cloud configurations at the same time.

It was in this context that the NAS technique known as Once-For-All (OFA) was presented as a solution with high efficiency and environmental sustainability characteristics~\cite{cai2019once}. 
The key idea was to train a single large network with a generic architecture and then fine-tune, truncate and adapt it for different hardware platforms and resource constraints, while keeping the initial macro-architecture fixed.
This goal is achieved through a training technique the authors called Progressive Shrinking (PS), which consists of many phases to progressively fine-tune smaller and smaller networks from within the supernet.
The learning process of the subnets is facilitated by the use of a knowledge distillation (KD) technique~\cite{hinton2015distilling}.
The OFA algorithm ends by returning the trained supernet, which can be easily adapted to a variety of different tasks and hardware with a further subnet search process.
Compared to other hardware-aware NAS techniques, this is considered to be one of the most environmentally friendly.

In this work, we present OFAv2, the extension of OFA with respect to the new design trends in the field of deep learning architecture design, with the aim of improving the supernet while preserving the climate sustainability properties.
In particular, we improve the OFAMobileNetV3 architecture by adding parallel blocks, additional dense skip connections, and implementing early exits.
To support the above architectural changes, the PS training algorithm has been expanded to the Extended Progressive Shrinking algorithm with the presence of two new elastic steps, i.e. Elastic Level and Elastic Height. The former is applied to support parallel networks, while the latter is applied when early exits are present.
Furthermore, we enrich the algorithm with a strategy to extract the teacher network from the supernet at the end of each PS phase. This approach is compared to the one proposed in OFA, which instead requires this network to be unchanged.

The rest of the article is structured as follows: 
The Section~\ref{sec: related_works} presents a summary of NAS hardware-aware techniques and summarises the approach taken in OFA, explaining how it works in order to have a reference point on which to base the rest of the concepts discussed in the article.
The Section~\ref{sec:method} presents the changes made to the OFAMobileNetV3 network from an architectural point of view, and the way in which the training algorithm was extended to support these changes.
The Section~\ref{sec:setup} describes the configurations used in the experiments.
Section~\ref{sec:results} reports the results obtained with the extended training algorithm for the new networks, comparing them with the basic networks and commenting on their quality.
Finally, Section~\ref{sec:conclusion} concludes the paper by summarising the main contributions and grouping observations on possible next steps.

\section{Related Works}\label{sec: related_works}

Neural Architecture Search (NAS) is a method for automating the design of neural network architectures. The goal of NAS is to find the optimal architecture for a given task, such as image classification or language translation, by searching through a large space of possible architectures. This is typically done using machine learning techniques such as reinforcement learning or evolutionary algorithms to guide the search process. The ultimate goal is to achieve better performance than architectures designed by human experts and to improve the efficiency of the design process.

\subsection{Hardware-Aware NAS}\label{subsec:hardware-aware NAS}

The first NAS techniques were aimed solely at finding the most accurate architecture possible, at the expense of all other metrics, such as the consumption and timing of such searches, or the required performance of the hardware that would exploit the neural network found~\cite{zoph2016neural, baker2016designing, xie2017genetic}.
More recently, some work has been proposed to optimise these searches by considering trade-offs in terms of timing and hardware capabilities.

Cai~\textit{et al.} developed ProxylessNAS, a NAS technique that optimises networks directly for the target task and hardware, achieving cutting-edge performance on image classification tasks for mobile-sized neural networks~\cite{cai2018proxylessnas}.
Tan~\textit{et al.} developed MnasNet and defined a novel search space tailored to the constraints of mobile devices, such as memory and computational budget. They also introduced a new search algorithm that incorporates a performance prediction model and a reinforcement learning-based controller to guide the search process~\cite{tan2019mnasnet}.
Other works have developed new NAS techniques by optimising a multi-objective problem to maximise accuracy by minimising the hardware load associated with training the networks, thereby reducing the overall search and training time of the optimal architecture~\cite{lomurno2021pareto, falanti2022popnasv2, falanti2022popnasv3}.
Wu~\textit{et al.} presented a new method for NAS called FBNet, which was designed to find efficient architectures for on-device inference. The authors proposed a differentiable method that uses a performance prediction model to predict the inference latency and energy consumption of different architectures~\cite{wu2019fbnet}.

\subsection{Once-For-All}\label{subsec:ofa}

Once-For-All (OFA) is a NAS technique entirely designed to realise high-performance neural networks that are easily adaptable to different hardware configurations, and to do so with the lowest possible power consumption~\cite{cai2019once}.
The idea is to search for a supernet that contains several possible subnets that are powerful and easily adaptable to different tasks and devices.
This approach is extremely advantageous in terms of time and CO$_2$ produced compared to other NAS techniques optimised for the same purpose.

Among the major macro architectures presented by the authors, the primary one, referred to as OFAMobileNetV3 (OFA\_MBV3), serves as the basis for this study and is outlined in Table~\ref{table:OFAmbv3}.
The network is based on the ``Inverted Residual Bottleneck"(IRB), a type of block based on depthwise separable convolutions introduced in MobileNetV2~\cite{sandler2018mobilenetv2} and further refined in MobileNetV3~\cite{howard2019searching}. 
By default, the network consists of 5 stages, each consisting of 4 blocks. When the number or size of feature maps from the input to the output of a block is changed, the residual connection cannot be used, and for this reason the IRB block is replaced by its sequential version called an ``Inverted Bottleneck" (IB).

\begin{table}[t]
\centering
\caption{Structure of OFA\_MBV3 (width multiplier 1.0), the base architecture by the authors from which we started. ``BN" represents the presence of Batch Normalization layers and ``SE" represents the presence of Squeeze and Excitation sub-blocks~\cite{hu2018squeeze} inside the IB and IRB blocks.}
    \begin{tabular}{@{} l lccccc @{}}
    \toprule
    \textbf{Stage}           & \textbf{Layer} & \textbf{Out Ch} & \textbf{Stride} & \textbf{Activation} & \textbf{BN} & \textbf{SE} \\
    \midrule
    \multirow{2}{*}{Head}    & conv3x3        & 16              & 2               & Hswish       & \checkmark           &           \\
                             & IRB            & 16              & 1               & ReLU         & \checkmark           &           \\ \midrule
    \multirow{4}{*}{Stage 1} & IB            & 24              & 2               & ReLU         & \checkmark           &           \\
                             & IRB            & 24              & 1               & ReLU         & \checkmark           &           \\
                             & IRB            & 24              & 1               & ReLU         & \checkmark           &           \\
                             & IRB            & 24              & 1               & ReLU         & \checkmark           &           \\ \midrule
    \multirow{4}{*}{Stage 2} & IB            & 40              & 2               & ReLU         & \checkmark           & \checkmark           \\
                             & IRB            & 40              & 1               & ReLU         & \checkmark           & \checkmark           \\
                             & IRB            & 40              & 1               & ReLU         & \checkmark           & \checkmark           \\
                             & IRB            & 40              & 1               & ReLU         & \checkmark           & \checkmark           \\ \midrule
    \multirow{4}{*}{Stage 3} & IB            & 80              & 2               & Hswish       & \checkmark           &           \\
                             & IRB            & 80              & 1               & Hswish       & \checkmark           &           \\
                             & IRB            & 80              & 1               & Hswish       & \checkmark           &           \\
                             & IRB            & 80              & 1               & Hswish       & \checkmark           &           \\ \midrule
    \multirow{4}{*}{Stage 4} & IB            & 112             & 1               & Hswish       & \checkmark           & \checkmark           \\
                             & IRB            & 112             & 1               & Hswish       & \checkmark           & \checkmark           \\
                             & IRB            & 112             & 1               & Hswish       & \checkmark           & \checkmark           \\
                             & IRB            & 112             & 1               & Hswish       & \checkmark           & \checkmark           \\ \midrule
    \multirow{4}{*}{Stage 5} & IB            & 160             & 2               & Hswish       & \checkmark           & \checkmark           \\
                             & IRB            & 160             & 1               & Hswish       & \checkmark           & \checkmark           \\
                             & IRB            & 160             & 1               & Hswish       & \checkmark           & \checkmark           \\
                             & IRB            & 160             & 1               & Hswish       & \checkmark           & \checkmark           \\ \midrule
    \multirow{4}{*}{Tail}    & conv1x1        & 960             & 1               & Hswish       & \checkmark           &           \\
                             & GAP            & 960             &                &             &            &            \\
                             & conv1x1        & 1280            & 1               & Hswish       &           &           \\
                             & Linear         & n classes       &                &             &           &           \\
    \bottomrule
    \end{tabular}
\label{table:OFAmbv3}
\end{table}

The main technique used by the authors to create optimal subntes within the supernet is called Progressive Shrinking (PS), and is the core of the OFA algorithm. The idea is to start with a large network, e.g. OFA\_MBV3, and gradually reduce the complexity of the subnets that can be activated by reducing the number of convolutional filters and the resolution of the input images. 
Thanks to this algorithm, the results shown by the authors demonstrate how OFA overcomes the ``multi-model forgetting'' problem~\cite{benyahia2019overcoming}, i.e. the problem that arises when weight sharing is used to sequentially train a number of neural architectures within the supernet, which generally leads to severe performance degradation.
The PS algorithm is divided into four ``elastic steps'', some of which consist of several phases.
Considering that each phase corresponds to a complete training run of the supernet, the elastic steps proposed by the authors are the following:

\begin{itemize}
    \item \textbf{Elastic Resolution (ER)}: 
    the size of the input images entering the network varies randomly for each batch with respect to the set of allowed values, i.e. [128, 160, 192, 224]. 
    This step involves a single phase and is active throughout the entire PS process.
    \item \textbf{Elastic Kernel Size (EKS)}: 
    the kernel size within the IRB/IB blocks is set to the maximum value chosen by the authors, i.e. 7. The weights associated with these convolution operations are computed throughout the training of the supernet. Next, the equivalent versions of the kernels of size 5 and 3 are computed using two small neural networks, with the aim of learning the matrix transformations that convert the kernels from size 7 to 5 and then from 5 to 3. This step  consists of a single phase.
    \item \textbf{Elastic Depth (ED)}: 
    the IRB blocks inside each stage of the supernet are deactivated sequentially, starting from the one closest to the output, allowing the training of subnets composed only of the first k blocks, with k less than or equal to N=4.
    Before this training step is activated, all phases use a k value fixed at 4.
    Once activated, Elastic Depth consists of two phases: in phase~1, the k value can be selected in [3,4], while in phase~2 the k value can be selected in [2,3,4].
    \item \textbf{Elastic Width (EW)}: 
    the width, i.e. the number of channels within the intermediate convolutions of the IRB/IB blocks, is set to the maximum value chosen by the authors, i.e. those in Table~\ref{table:OFAmbv3} multiplied by a factor f=6.
    Once activated, this elastic step consists of two phases: in phase~1, f can be chosen in [4,6], while in phase~2, f values can be chosen in [3,4,6].
    Whenever values of f less than 6 are selected for a block, the algorithm orders the channel weights of the intermediate layers hierarchically according to their L1 norm. It then calculates the number of channels to be used in the current phase and fills them with the best weights taken from the ranking.
\end{itemize}

At the beginning of the algorithm, the maximal network is defined, i.e. the OFA\_MBV3 architecture with all PS parameters set to their maximum value. 
The training steps and phases of PS are then executed in the order described above
All values unlocked by a given phase for a given elastic step remain available for selection by all subsequent training phases, allowing smaller and smaller networks to be added to the sampling space.
For each batch of images, a certain number of subnets are sampled from the current sample space, i.e. the set of the possible activatable subnets for that particular PS phase. Their gradients are cumulated, and then a single weight update step is performed.
The progressive addition of smaller subnets to the sampling space, whose weights are also derived from the tuning of progressively smaller networks, minimises the chance of interference between subnets, thus limiting the impact of the multi-model forgetting problem.

Throughout the training process, the maximal network is used as the teacher network to perform Knowledge Distillation (KD)~\cite{hinton2015distilling}. 
KD is a technique for transferring knowledge from a large, pre-trained network (called the teacher network, in this case the maximal network) to a smaller network (called the student network, in this case an active subnet). The idea is to use the output of the teacher network as soft targets for the student network during training. This allows the student network to learn from the predictions of the teacher network, resulting in a smaller network with similar performance to the larger teacher network.
The OFA algorithm uses a form of KD where the loss of the student network is computed as the sum of its cross-entropy loss and the loss computed on the soft labels of the teacher network, this second term being weighted by a KD-ratio factor.
For more information on the OFA technique, please refer to the original article.

\section{Method}
\label{sec:method}
This section presents OFAv2, describing how to extend the OFA\_MBV3 architecture by adding dense skip connections, early exits, and parallel blocks.
These changes increase the number and possible types of subnets that can be expressed within the supernet.
Then, the training technique called Extended Progressive Shrinking (EPS) is introduced, which extends the PS technique by adding the Elastic Level and Elastic Height steps.
Finally, the Ensembled Knowledge Distillation technique used for training subnets is presented.

\subsection{Architectural modifications}
\label{subsec:ofa_networks}

\begin{figure}[t]
\centering
\includegraphics[scale=0.10]{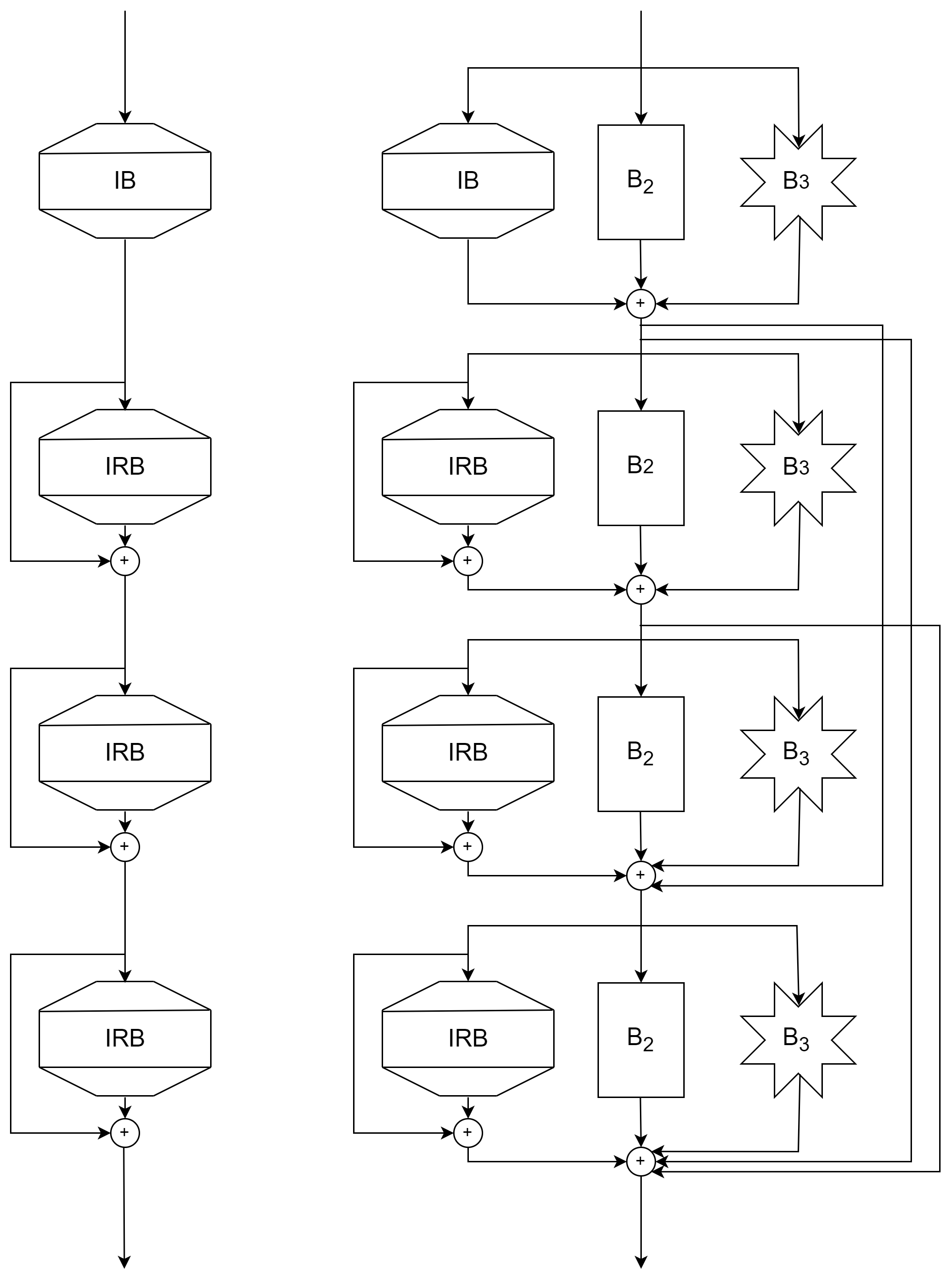}
\caption{On the left, the structure of a stage in the baseline network. On the right, the structure of a stage for a network endowed with both parallel blocks and dense skip connection}
\label{fig:b_vs_dp}
\end{figure}
\begin{figure}[t]
\centering
\includegraphics[scale=0.1575]{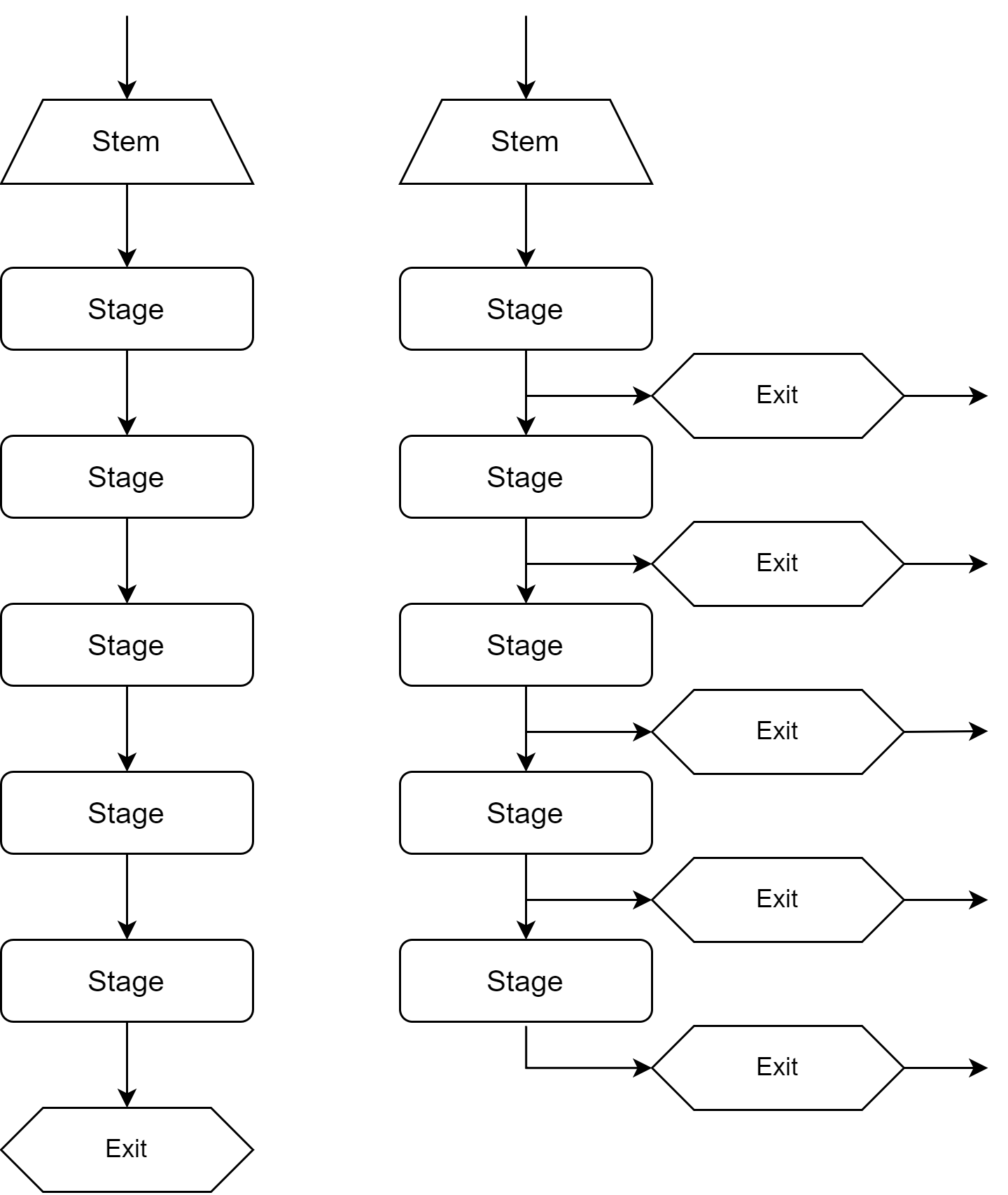}
\caption{On the left, the staged structure of the baseline network. On the right, the staged structure of a network enriched with early exits}
\label{fig:se_vs_me}
\end{figure}

The work focused on 3 main architectural changes with respect to OFA\_MBV3, which can be considered as direct extensions of the original structure. 
In detail, the modifications introduced in our work are:
\begin{itemize}
    
    \item \textbf{Dense Skip Connections (D)}: 
    the supernet is extended with a dense scheme of skip connections between the blocks belonging to the same stage.
    Compared to the original OFA\_MBV3 structure, where the output of each block is a residual obtained through the summation with the input, we also add the output of the previous blocks in the stage.
    Two-block skips are performed by the additional skip connections, to not duplicate the ones already present in the IRB blocks.
    
    \item\textbf{Early Exits (EE)}: 
    the Supernet is no longer characterised by a single output, but is enriched with several intermediate outputs, called early exits.
    In particular, one exit is placed after each network stage, for a total of 5 exits.
    The idea behind early exits is that intermediate outputs could have comparable or even better performance than the final one, making it possible to consider the use of computationally lighter models~\cite{sarti2023anticipate}.
    Indeed, given the constraints on the search process, the ability to exit the network early could lead to significant savings in terms of parameters, flops and latency.
    The use of early exits improves the performance of the intermediate stages, facilitating the a posteriori search for the best subnets to deploy on resource-constrained devices.

    \item \textbf{Parallel Blocks (P)}: 
    this extension consists of the implementation of two new blocks that are executed in parallel with the IRBs/IBs.
    We refer to as \emph{level} the set of blocks executed in parallel at a given depth of a stage.
    This modification increases the variety of architectures that are searchable by the algorithm, since subpaths consisting of one or more blocks can be selected at each level.
    The first additional block consists of a sequence composed of one pointwise convolutional layer, a batch normalisation layer, and an activation function corresponding to the one used in the IRB/IB. 
    The goal of this block is to mix the features in the channel dimension.
    The second block focuses instead on fast transformations, providing a much more lighter alternative compared to the IRB block.
    This block consists only of a batch normalisation layer followed by an activation function  corresponding to the one used in the IRB/IB, preceded by additional operators in case the output tensor needs to be reshaped to match those of the other blocks in that level.
    In detail, a max-pooling operation can be applied to reduce the spatial dimension of the feature maps, and, similarly, a 1x1 convolution can be used to increase the number of feature maps.
    The choice of such simple additional blocks was dictated by the desire to limit as much as possible the inevitable increase in the number of parameters and flops.

\end{itemize}

\begin{figure}[t]
\centering
\includegraphics[scale=0.16]{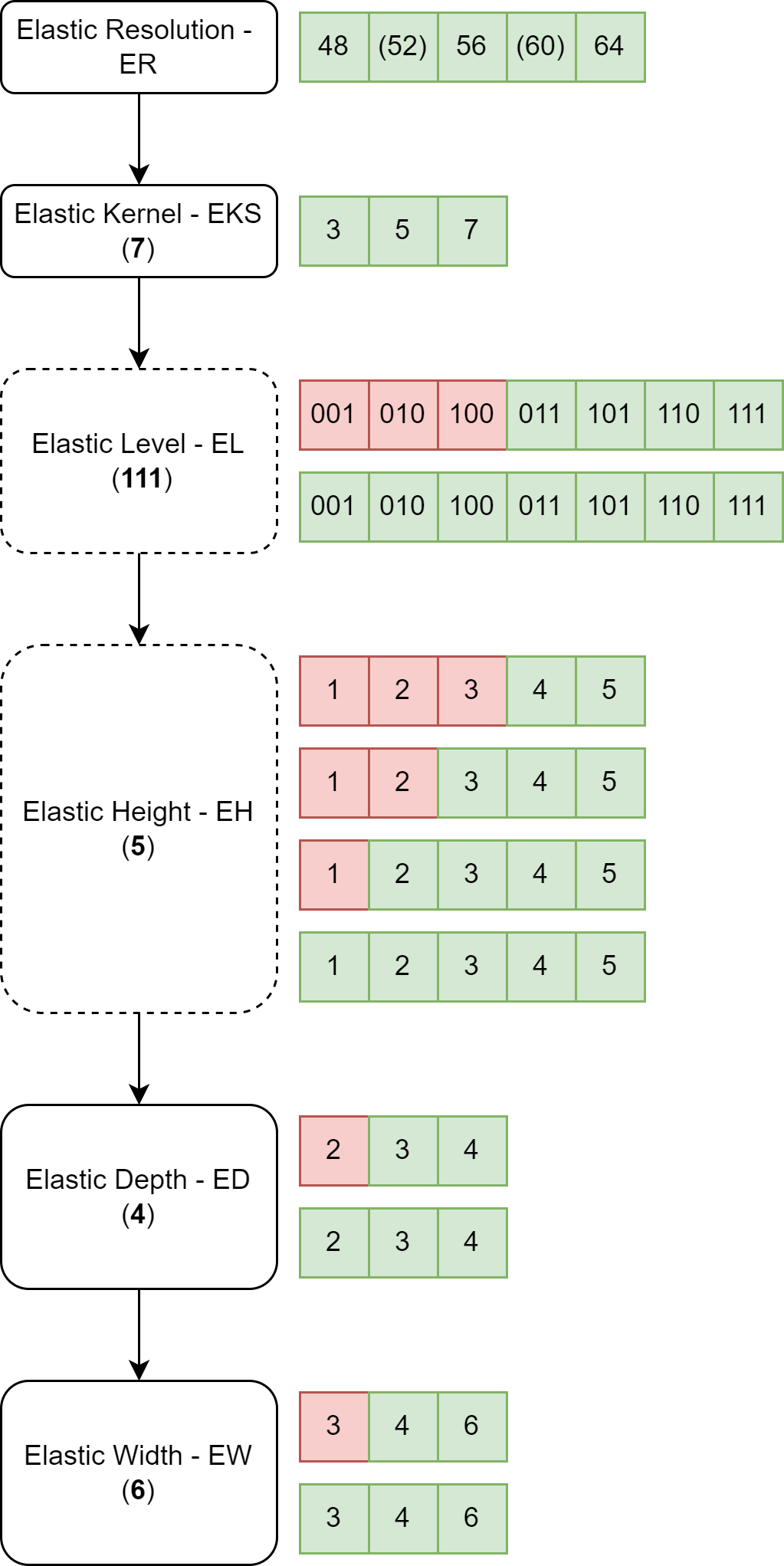}
\caption{
Extended Progressive Shrinking: each step can be composed of multiple phases, allowing the training algorithm to progressively activate smaller sub-networks.
The figure illustrates the sequential steps performed by the training procedure, accompanied on the right by the allowed set of values in our experiments for each phase of that step.
The available options for a given phase are shown in green, while the unavailable ones are reported in red.
When a step has not been reached yet, the value for the corresponding modifier is set to the maximum one, reported under the name of the step.
Dashed steps are performed only on supernets supporting the architectural modifications touched by the step,
namely parallel blocks for the Elastic Level, and early exits for the Elastic Height.
}
\label{fig:extended_ps}
\end{figure}

The names and characteristics of the networks tested are summarised in Table~\ref{table:ofa_networks_summary}.
We adopt the following naming convention for new configurations, consisting of two prefixes.
The first prefix can be either SE or EE, representing single-exit and early-exit networks respectively.
The second prefix refers to inter-stage modifications: D indicates the presence of dense skip connections, P that of parallel blocks, and B (base) the absence of these last two modifications.
\begin{table}[t]
\centering
\caption{Names of networks and corresponding structural changes supported}

\renewcommand*{\arraystretch}{1.25}
\begin{tabularx}{\linewidth}{@{} l YYc @{}}
    \toprule
    \textbf{Network Name} & \textbf{Early Exits} & \textbf{Dense Skip} & \textbf{Parallel Blocks} \\
    \midrule
    OFA\_MBV3 (baseline)  &  &  &   \\ \midrule
    SE\_D\_OFA\_MBV3   &      & \checkmark         &        \\ \midrule
    SE\_P\_OFA\_MBV3   &      &          & \checkmark       \\\midrule
    SE\_DP\_OFA\_MBV3  &      & \checkmark         & \checkmark       \\ \midrule
    EE\_B\_OFA\_MBV3   & \checkmark     &          &        \\ \midrule
    EE\_D\_OFA\_MBV3   & \checkmark     & \checkmark         &        \\ \midrule
    EE\_P\_OFA\_MBV3   & \checkmark     &          & \checkmark       \\ \midrule
    EE\_DP\_OFA\_MBV3  & \checkmark     & \checkmark         & \checkmark       \\
    \bottomrule
\end{tabularx}
\label{table:ofa_networks_summary}
\end{table}

Figure~\ref{fig:b_vs_dp} and Figure~\ref{fig:se_vs_me} graphically represent the architectural modifications made at the stage and network levels respectively.

\subsection{Extended Progressive Shrinking}
\label{subsec:ext_ps}
The Extended Progressive Shrinking (EPS) technique is a modified version of the PS proposed in the OFA algorithm. Its objective is to take into account the proposed architectural changes made to the supernet, allowing the training process to take into account the newly introduced subnetworks.
Thus, two additional steps are designed, following the same philosophy described in the PS, i.e. training the largest architectures first and sampling smaller subnetworks progressively:

\begin{itemize}
\item \textbf{Elastic Level (EL)}: 
this elastic step is only performed if the supernet under consideration contains parallel blocks.
When used, subnets can be extracted from the supernet so that at each level one or more blocks can be selected to be part of the subnet.
An integer value is assigned to each level, the corresponding 3-bit one-hot coding representing the activation state of the 3 blocks that make up the level.
Before EL is activated, all previous training phases use a value fixed to 7 (or 111), which means that all 3 blocks in the levels are trained together.
Once active, this elastic step consists of two phases: phase~1 allows EL to train levels by selecting pairs of blocks, phase~2 further extends the choices to allow training of just one of the three blocks.

\begin{table}[t]
\centering
\caption{Extended Progressive Shrinking phases hyperparameters}
    \begin{tabularx}{\linewidth}{@{} l  @{\hskip 3em}  YYYc @{}}
    \toprule
    \textbf{Phase}       & \textbf{LR}   & \textbf{Epochs}    & \textbf{Warmup Ep.}     & \textbf{\#Subnets}   \\
    \midrule
    Full    & $1.0\cdot10^{-3}$     & 180       & 0             & 0         \\
    EKS      & $3.0\cdot10^{-2}$     & 120       & 5             & 1         \\
    EL1     & $2.5\cdot10^{-3}$     & 25        & 0             & 2         \\
    EL2     & $7.5\cdot10^{-3}$     & 120       & 5             & 2         \\
    EH1     & $2.5\cdot10^{-3}$     & 25        & 0             & 2         \\
    EH2     & $7.5\cdot10^{-3}$     & 60        & 5             & 2         \\
    EH3     & $1.0\cdot10^{-2}$     & 90        & 5             & 2         \\
    EH4     & $3.0\cdot10^{-2}$     & 120       & 5             & 2         \\
    ED1      & $2.5\cdot10^{-3}$     & 25        & 0             & 2         \\
    ED2      & $7.5\cdot10^{-3}$     & 120       & 5             & 2         \\
    EW1      & $2.5\cdot10^{-3}$     & 25        & 0             & 4         \\
    EW2      & $7.5\cdot10^{-3}$     & 120       & 5             & 4         \\
    \bottomrule
    \end{tabularx}
\label{tab: eps_hyperpar}
\end{table}

\item \textbf{Elastic Height (EH)}: 
This elastic step is only performed if the considered supernet is enriched with early exits.
When used, subnets can be extracted from the supernet so that the subnets are single-exit networks terminating at the N\textsuperscript{th} exit.
Prior to its activation, all previous training phases use a value of N fixed at 5, which means that subnets always use the last exit. 
Once active, this elastic step consists of four phases: phase~1 allows N to be chosen in [4,5], phase~2 extends the choice to [3,4,5], phase~3 to [2,3,4,5], and finally phase~4 to [1,2,3,4,5].
\end{itemize}

The EPS pipeline is shown in Figure~\ref{fig:extended_ps}.
When both steps are required, the Elastic Level step is performed before the Elastic Height step and after the Elastic Kernel Size step.
The maximal network when these two changes are implemented is defined as the one with all values for each elastic step fixed to its maximum, and all 5 exits active.
While the original PS implementation uses as teacher network the version with the latest weights saved just after the first Elastic Resolution step, in EPS algorithm the teacher network, i.e. the maximal network, is updated at the beginning of each phase by extracting its weights and configuration the from the supernet obtained after the previous phase.

\subsection{Training Early-Exit Networks}
During the first training step of supernets with early exits, the main goal is to maximise the performance of such networks with all exits active.
To optimise the training and evaluation of such networks, we follow the early exits ensemble approach proposed in the ``Anticipate, Ensemble and Prune"(AEP) method following DESC strategy~\cite{sarti2023anticipate}.
This technique consists of computing the loss of the network as the weighted sum of the losses of the exits, while the accuracy of the network is computed as the argmax of the weighted sum of the early exit outputs.
The post-processing pruning step is not applied because the whole network is needed as a teacher by the following EPS training steps.
The presence of early exits also requires the adoption of a new form of KD.
For this case, we devise a form of ensembled KD (ENS-KD) where for each selected subnet, regardless of the chosen exit, the algorithm is performed with respect to the ensemble of all teacher network exits.
Such ensemble is generated following the technique described in AEP, using the DESC weight assignment strategy.

\section{Experiments Setup}\label{sec:setup}

To test the quality of the refinements introduced in this work, we conducted experiments for each supernet configuration using the same setup to ensure a fair comparison of the results.
In this section, we describe how the experiments were carried out, defining the dataset, the set of hyperparameters, and the setup for the EPS procedure.

\subsection{Supernet configurations} \label{subsec:experiments}

The experiments were run for all the eight supernet configurations previously shown in Table~\ref{table:ofa_networks_summary}, to compare their results and isolate the impact of each modification on the final performance.
In our study, the original OFA\_MBV3 is the baseline network for studying the effect of our architectural changes.
Each supernet configuration underwent the training process four times, as we train them for each combination of width multiplier (WM, i.e. 1.0 or 1.2) and teacher network strategy (i.e. fixed to the one obtained at the first step or being progressively updated).
The width multiplier is a multiplicative hyperparameter introduced in~\cite{howard2017mobilenets}, which linearly scales the number of feature maps generated by all blocks in the network.

\subsection{Dataset}

While the original OFA architecture was trained on ImageNet~\cite{deng2009imagenet}, in this work the OFAv2 algorithm was tested over the Tiny ImageNet dataset~\cite{le2015tiny},
which is a subset of ImageNet consisting of \numprint{100000} training images (15\% of which were used for validation) and \numprint{10000} test images, downsampled to a 64x64 resolution.
Due to the smaller size, both in terms of number of images and image size, it is possible to perform the experiments in a reasonable amount of time, making them reproducible without significant hardware investment.
The OFA v2 algorithm is implemented using PyTorch 1.12.1, and each experiment was run on a NVIDIA QUADRO RTX 6000, taking an average of one and a half days to complete. In comparison, the original OFA training on ImageNet was reported to take one and a half days on a cluster of 32 NVIDIA V100 GPUs.

\subsection{Hyperparameters}

For the training procedure, the SGD optimizerwas adopted with momentum equal to $0.9$ and a weight decay parameter set to $3\cdot10^{-5}$. 
The learning rate was adjusted at each iteration using a Cosine Annealing scheduler.
The batch size was set to 200.
The hyperparameters used in the EPS training phases are shown in Table~\ref{tab: eps_hyperpar}.
Elastic Resolution was run within the sizes in the set [48, 56, 64], using the original image size as the upper limit, as done by OFA for ImageNet.

\section{Results and Discussion}\label{sec:results}


\begin{table*}[p]
\caption{
For all the networks using the width multiplier WM=1.0, for each of the Extended Progressive Shrinking step (after the last phase has been executed), the table shows the results obtained by the tested subnets in terms of both average accuracy (``avg") and the accuracy of the best among them (``best").
For each network the first row represents the results obtained when the teacher network is fixed to the one obtained at the first step of the process, while the second row shows the results obtained when the teacher network is extracted from the supernet obtained as a result of the previous phase of the algorithm.
The symbol ``*" represents our new EPS steps, not present in OFA.
The value ``X" indicates that for a specific network, that PS step was not necessary and therefore it was skipped.
The best results for each step are highlighted in bold.
The best overall results are highlighted with underlining.
}
\begin{tabularx}{\linewidth}{|l||YY|YY|YY|YY|YY|YY|}
\toprule
\multirow{2}{*}{\textbf{NETWORK (WM = 1.0)}} & \multicolumn{2}{c|}{\textbf{RESOLUTION}} & \multicolumn{2}{c|}{\textbf{KERNEL SIZE}} & \multicolumn{2}{c|}{\textbf{LEVEL *}} & \multicolumn{2}{c|}{\textbf{HEIGHT *}} & \multicolumn{2}{c|}{\textbf{DEPTH}} & \multicolumn{2}{c|}{\textbf{WIDTH}} \\
 & avg & best & avg & best & avg & best & avg & best & avg & best & avg & best \\
\midrule
\multirow{2}{*}{OFA\_MBV3 (baseline)} & 27.28 & 28.14 & 37.46 & 38.97 & X & X & X & X & 37.67 & 39.35 & 38.11 & 39.83 \\
 & 27.03 & 28.02 & 37.90 & 39.38 & X & X & X & X & 39.45 & 41.46 & 40.03 & 42.23 \\ \midrule
\multirow{2}{*}{SE\_D\_OFA\_MBV3} & 28.49 & 30.03 & 38.90 & 40.49 & X & X & X & X & 38.85 & 40.91 & 38.77 & 40.93 \\
 & 29.50 & 30.56 & 39.73 & \textbf{41.60} & X & X & X & X & 40.37 & 42.34 & 40.83 & 43.07 \\ \midrule
\multirow{2}{*}{EE\_B\_OFA\_MBV3 } & 18.08 & 19.01 & 37.08 & 38.51 & X & X & 41.49 & 47.17 & 40.55 & 48.24 & 40.68 & 48.80 \\
 & 17.81 & 18.76 & 36.94 & 38.51 & X & X & \textbf{43.93} & 48.76 & \textbf{43.47} & 50.00 & 44.18 & 51.64 \\ \midrule
\multirow{2}{*}{EE\_D\_OFA\_MBV3 } & 17.68 & 18.36 & 38.35 & 40.00 & X & X & 41.04 & 46.37 & 40.11 & 47.63 & 40.30 & 48.53 \\
 & 17.96 & 18.59 & 38.11 & 39.76 & X & X & 43.48 & \textbf{48.82} & 43.31 & \textbf{50.62} & \underline{\textbf{44.22}} & \underline{\textbf{51.90}} \\ \midrule
\multirow{2}{*}{SE\_P\_OFA\_MBV3} & 28.21 & 29.41 & 38.68 & 40.15 & 23.80 & 39.89 & X & X & 24.73 & 38.89 & 24.51 & 37.93 \\
 & 27.98 & 28.95 & 37.99 & 39.36 & 23.65 & 39.35 & X & X & 24.03 & 37.92 & 23.42 & 36.17 \\ \midrule
\multirow{2}{*}{SE\_DP\_OFA\_MBV3} & 30.32 & 31.40 & 39.71 & 41.10 & 25.47 & 41.26 & X & X & 26.28 & 40.39 & 26.09 & 39.13 \\
 & \textbf{30.39} & \textbf{31.98} & \textbf{39.75} & 41.29 & 24.84 & 41.00 & X & X & 24.46 & 38.78 & 23.27 & 36.55 \\ \midrule
\multirow{2}{*}{EE\_P\_OFA\_MBV3 } & 16.06 & 16.74 & 37.19 & 38.73 & 23.59 & 39.36 & 26.97 & 44.33 & 27.33 & 43.85 & 27.96 & 44.12
 \\
 & 16.49 & 17.47 & 37.02 & 38.53 & 24.17 & 40.79 & 27.49 & 45.73 & 28.12 & 45.23 & 29.18	& 46.06\\ \midrule
\multirow{2}{*}{EE\_DP\_OFA\_MBV3 } & 15.96 & 16.96 & 37.89 & 39.36 & 25.10 & 40.84 & 27.58 & 45.73 & 27.72 & 44.89 & 28.10 &	44.67
 \\
 & 16.19 & 17.07 & 38.03 & 39.63 & \textbf{25.38} & \textbf{42.42} & 28.37 & 46.74 & 28.95 & 45.89 & 30.20 &	46.24 \\ \bottomrule
\end{tabularx}
\label{table:res_1.0}
\end{table*}
\begin{table*}[p]
\caption{
For all the networks using the width multiplier WM=1.2, for each of the Extended Progressive Shrinking step (after the last phase has been executed), the table shows the results obtained by the tested subnets in terms of both average accuracy (``avg") and the accuracy of the best among them (``best").
For each network the first row represents the results obtained when the teacher network is fixed to the one obtained at the first step of the process, while the second row shows the results obtained when the teacher network is extracted from the supernet obtained as a result of the previous phase of the algorithm.
The symbol ``*" represents our new EPS steps, not present in OFA.
The value ``X" indicates that for a specific network, that PS step was not necessary and therefore it was skipped.
The best results for each step are highlighted in bold.
The best overall results are highlighted with underlining.
}
\begin{tabularx}{\linewidth}{|l||YY|YY|YY|YY|YY|YY|}
\toprule
\multirow{2}{*}{\textbf{NETWORK (WM = 1.2)}} & \multicolumn{2}{c|}{\textbf{RESOLUTION}} & \multicolumn{2}{c|}{\textbf{KERNEL SIZE}} & \multicolumn{2}{c|}{\textbf{LEVEL *}} & \multicolumn{2}{c|}{\textbf{HEIGHT *}} & \multicolumn{2}{c|}{\textbf{DEPTH}} & \multicolumn{2}{c|}{\textbf{WIDTH}} \\
 & avg & best & avg & best & avg & best & avg & best & avg & best & avg & best \\
\midrule
\multirow{2}{*}{OFA\_MBV3 (baseline)} & 27.48 & 28.37 & 38.38 & 39.75 & X & X & X & X & 39.45 & 41.46 & 40.03 & 42.23 \\
 & 27.74 & 28.66 & 39.54 & 40.62 & X & X & X & X & 40.54 & 41.82 & 41.15 & 42.71 \\ \midrule
\multirow{2}{*}{SE\_D\_OFA\_MBV3} & 30.05 & 31.38 & 40.48 & 41.98 & X & X & X & X & 40.23 & 42.17 & 40.32 & 42.44 \\
 & 29.90 & 30.82 & 41.55 & \textbf{43.11} & X & X & X & X & 42.45 & 44.20 & 43.20 & 45.44 \\ \midrule
\multirow{2}{*}{EE\_B\_OFA\_MBV3 } & 21.32 & 22.24 & 39.13 & 40.45 & X & X & 43.42 & 47.79 & 42.38 & 48.40 & 42.64 & 49.49 \\
 & 22.07 & 23.38 & 39.51 & 40.97 & X & X & 46.26 & 50.52 & 46.09 & 51.83 & 47.13 & 53.37 \\ \midrule
\multirow{2}{*}{EE\_D\_OFA\_MBV3 } & 20.86 & 21.98 & 39.77 & 41.46 & X & X & 43.24 & 47.47 & 42.56 & 48.47 & 42.79 & 49.93 \\
 & 21.06 & 21.87 & 40.12 & 41.91 & X & X & \textbf{46.70} & \textbf{51.07} & \textbf{46.62} & \textbf{52.47} & \underline{\textbf{47.46}} & \underline{\textbf{53.76}} \\ \midrule
\multirow{2}{*}{SE\_P\_OFA\_MBV3} & 29.19 & 30.54 & 39.75 & 40.95 & 24.75 & 40.98 & X & X & 26.11 & 40.66 & 26.38 & 39.87 \\
 & 28.37 & 29.65 & 39.01 & 40.61 & 24.57 & 40.23 & X & X & 25.35 & 39.01 & 24.87 & 37.35 \\ \midrule
\multirow{2}{*}{SE\_DP\_OFA\_MBV3} & 31.12 & 32.38 & 40.46 & 42.11 & \textbf{26.38} & 41.77 & X & X & 27.28 & 41.16 & 27.57 & 40.14 \\
 & \textbf{31.31} & \textbf{32.97} & \textbf{40.92} & 42.53 & 26.19 & 41.93 & X & X & 26.45 & 39.73 & 25.97 & 38.02 \\ \midrule
\multirow{2}{*}{EE\_P\_OFA\_MBV3 } & 19.39 & 20.28 & 38.48 & 40.08 & 24.52 & 40.87 & 28.90 & 46.18 & 29.19 & 45.65 & 29.89 & 46.01
 \\
 & 19.47 & 20.59 & 38.79 & 40.36 & 24.89 & 41.55 & 29.73 & 48.12 & 30.71 & 47.84 & 32.26 &	48.44 \\ \midrule
\multirow{2}{*}{EE\_DP\_OFA\_MBV3 } & 18.71 & 18.99 & 38.38 & 40.24 & 25.89 & 42.03 & 29.51 & 46.88 & 30.03 & 46.60 & 30.28 &	46.74
 \\
 & 18.68 & 19.26 & 38.33 & 39.89 & 26.23 & \textbf{42.52} & 30.27 & 48.43 & 31.83 & 48.93 & 32.76	& 48.62\\ \bottomrule
\end{tabularx}
\label{table:res_1.2}
\end{table*}

As mentioned in Section~\ref{subsec:experiments}, the experiments have been carried out with respect to two different weight multiplier configurations, i.e. 1.0 and 1.2. The results are summarized in Table~\ref{table:res_1.0} and Table~\ref{table:res_1.2} respectively.
Accuracy scores over the test set are calculated at each phase of the proposed EPS algorithm.
For each of these phases, all possible combinations of configuration parameters are sampled, taking into account all previously unlocked phases.
As an example, the testing of EE\_B\_OFA\_MBV3 in phase EH2 (second phase of Elastic Height) would activate all subnets obtained by choosing resolutions in [48,56,64], kernel size in [3,5,7], network height in [3,4,5], stage depth in [4] and expansion-ratio in [6], for a total 27 subnets.
For each of the steps, the average accuracy values obtained by the subnets (avg) and the performance of the best subnet sampled (best) are shown.
Finally, for each configuration, the two approaches to teacher network selection are compared, with the results obtained using the technique presented in OFA on the top row and those obtained using our approach on the bottom row.

Looking at the results of the first step, characterised by the application of Elastic Resolution alone, it can be seen that the best configuration, both as an average and as the best subnet, turns out to be the one with single exit, dense skip connections and parallel blocks.
This result shows how it is possible to better generalise the information contained in images of different sizes by having a more varied set of blocks and a greater number of connections.
In this step, the teacher network that will be used in subsequent steps is produced (constant for OFA's approach, progressively updated in ours).
These results are confirmed regardless of the width multiplier considered.

The Elastic Kernel Size step represents an overall improvement for each configuration, especially for the models enriched with early exits, which gain approximately 20\% accuracy both on average and as a better subnet performance regardless of the width multiplier considered.
The best average configuration is still the one with a single exit, dense skip connections, parallel blocks, by applying a progressively updated teacher network.
The best subnet is the one with a single exit and dense skip connections, again using a progressively updated teacher network.

The Elastic Level step we introduced is applied to all configurations having parallel blocks.
It is possible to observe how in this case the configuration with the best accuracy is EE\_DP\_OFA\_MBV3, i.e. the one with all the modifications we made to the OFA\_MBV3 network. It is important to note that already at this EPS step the best subnet found is better than the overall best performing model obtained through OFA algorithm with fixed teacher network, regardless of the width multiplier considered.
Regarding the average performance within this stage, it can be observed how the choice of very light blocks leads to subnets that are very heterogeneous among themselves and with a poor performing majority.

The Elastic Height step we introduced is applied to all configurations with early exits.
From this step forward, it can be observed in both width multiplier configurations how single-block architectures manage to benefit better from EPS than those with parallel blocks.
The use of the dynamic teacher network begins to prove particularly beneficial in terms of average and best accuracy.
During the Elastic Depth and Elastic Width steps, applied to all possible architecture configurations, the considerations made regarding the Elastic Height step continue to hold true.
Thus, looking at the final result of the OFAv2 algorithm, it can be said that this represents a substantial improvement over the baseline represented by OFA.

In particular, looking at Table~\ref{table:res_1.0}, it can be seen that the EE\_D\_OFA\_MBV3 configuration gains an accuracy value of 12.07\% compared to the reference OFA algorithm. Looking at Table~\ref{table:res_1.2}, the same behaviour can be seen with a deviation of 11.53\%. It is also interesting to note that the progressive teacher updating technique alone results in an improvement in the model better than 3\% accuracy, regardless of the width multiplier.
Finally, it can be seen that the average performance of OFAv2 is better than that achieved by the best subnet obtained through OFA.

Concerning the width multiplier, the results obtained confirm how OFAv2 manages to scale up very well with the increase in the number of filters used, gaining almost 2\% accuracy with this simple parameter change.
From an overall perspective, the results show that using the defined parallel blocks does not have a beneficial effect on the final performance obtained on the whole training procedure.
This may have depended on the type of dataset rather than the type of task, and could therefore have opposite implications in other configurations.
On the contrary, the addition of early exits produces noticeable improvements over the baseline.

Interestingly, networks with early exits perform worse than their single-exit counterparts in the initial training step, a rather odd behaviour compared to the one showed in~\cite{sarti2023anticipate}, where early-exit networks clearly outperformed single-exit networks.
This behaviour is most likely to be attributed to the use of Elastic Resolution.
Single-exit networks are more suited for steadily optimizing the weights for various image sizes, while updating multiple exits to support all image sizes appears to be a difficult objective to achieve.
Anyway, from the Elastic Kernel Size step onward, networks with early exits close the gap and overcome the performances of single-exit networks, providing a significant accuracy improvement at the end of the training.
Adding dense skip connections to the base OFA\_MBV3 supernet provides a marginal improvement on the final architectures.
Even if the performance boost is minimal compared to the one provided by the early exits, their improvements stack, making it possible to combine these two techniques to maximize the subnets performance.

Finally, the use of progressively updated teacher networks, revised with the best configuration of weights after each phase of the algorithm, induces significant, consistent improvement in most of the EPS steps and network configurations.
In this way, the student networks are able to learn better from the more accurate teacher network, but they also improve the teacher weights shared with these subnets, creating a mutual learning benefit.
These results indicate that the applied architectural changes are effective for boosting the performance of all subnets contained in the supernet, which reach on average a better accuracy than the best subnet from in the baseline network.

\section{Conclusion}\label{sec:conclusion}

In this paper, we present OFAv2, the extended version of the Once-For-All algorithm, which enriches the supernet design with early exits, dense skip connections and parallel blocks.
The new architectural patterns are applied to the original OFAMobileNetv3 network, evaluating the different configurations on the Tiny ImageNet dataset.
To support the training of the new subnets, the Extended Progressive Shrinking algorithm extends its counterpart presented in the Once-For-All paradigm with two new steps, namely the Elastic Level and Elastic Height steps.
The Elastic Level step takes care of progressively sampling a smaller number of parallel blocks based on the training phase, while The Elastic Level can select different exits among those added in our work.

When trained with Extended Progressive Shrinking, early exits and dense skip connections provide a significant increase in accuracy over the original OFAMobileNetV3 architecture.
The network configurations enriched with early exits directly improve the performance of the intermediate layers of the supernet, making it easier to search for optimal subnets.
Furthermore, the progressive extraction of an updated version of the teacher network during the Extended Progressive Shrinking algorithm, instead of using the one obtained at the first step, contributes to a further improvement of the final performance.

The aim of this work was to improve one of the most cutting-edge hardware-aware NAS techniques by enabling it to achieve even better results, while remaining in an eco-friendly context suitable for developing neural networks for different hardware configurations.
The next steps include the study of new parallel blocks and the extension to different tasks and domains.
Follow-up work on an ex-post search process to support the new architectures will most likely lead to further performance improvements.

\section*{Acknowledgment}
The European Commission has partially funded this work under the H2020 grant N. 101016577 AI-SPRINT: AI in Secure Privacy-pReserving computINg conTinuum. This work has also been supported by the FAIR (Future Artificial Intelligence Research) project, funded by the NextGenerationEU program within the PNRR-PE-AI scheme (M4C2, investment 1.3, line on Artificial Intelligence).

\bibliographystyle{IEEEtran}
\bibliography{bibliography}

\begin{thebibliography}{10}
\providecommand{\url}[1]{#1}
\csname url@samestyle\endcsname
\providecommand{\newblock}{\relax}
\providecommand{\bibinfo}[2]{#2}
\providecommand{\BIBentrySTDinterwordspacing}{\spaceskip=0pt\relax}
\providecommand{\BIBentryALTinterwordstretchfactor}{4}
\providecommand{\BIBentryALTinterwordspacing}{\spaceskip=\fontdimen2\font plus
\BIBentryALTinterwordstretchfactor\fontdimen3\font minus
  \fontdimen4\font\relax}
\providecommand{\BIBforeignlanguage}[2]{{%
\expandafter\ifx\csname l@#1\endcsname\relax
\typeout{** WARNING: IEEEtran.bst: No hyphenation pattern has been}%
\typeout{** loaded for the language `#1'. Using the pattern for}%
\typeout{** the default language instead.}%
\else
\language=\csname l@#1\endcsname
\fi
#2}}
\providecommand{\BIBdecl}{\relax}
\BIBdecl

\bibitem{cai2019once}
H.~Cai, C.~Gan, T.~Wang, Z.~Zhang, and S.~Han, ``Once-for-all: Train one
  network and specialize it for efficient deployment,'' \emph{arXiv preprint
  arXiv:1908.09791}, 2019.

\bibitem{hinton2015distilling}
G.~Hinton, O.~Vinyals, J.~Dean \emph{et~al.}, ``Distilling the knowledge in a
  neural network,'' \emph{arXiv preprint arXiv:1503.02531}, vol.~2, no.~7,
  2015.

\bibitem{zoph2016neural}
B.~Zoph and Q.~V. Le, ``Neural architecture search with reinforcement
  learning,'' \emph{arXiv preprint arXiv:1611.01578}, 2016.

\bibitem{baker2016designing}
B.~Baker, O.~Gupta, N.~Naik, and R.~Raskar, ``Designing neural network
  architectures using reinforcement learning,'' \emph{arXiv preprint
  arXiv:1611.02167}, 2016.

\bibitem{xie2017genetic}
L.~Xie and A.~Yuille, ``Genetic cnn,'' in \emph{Proceedings of the IEEE
  international conference on computer vision}, 2017, pp. 1379--1388.

\bibitem{cai2018proxylessnas}
H.~Cai, L.~Zhu, and S.~Han, ``Proxylessnas: Direct neural architecture search
  on target task and hardware,'' \emph{arXiv preprint arXiv:1812.00332}, 2018.

\bibitem{tan2019mnasnet}
M.~Tan, B.~Chen, R.~Pang, V.~Vasudevan, M.~Sandler, A.~Howard, and Q.~V. Le,
  ``Mnasnet: Platform-aware neural architecture search for mobile,'' in
  \emph{Proceedings of the IEEE/CVF Conference on Computer Vision and Pattern
  Recognition}, 2019, pp. 2820--2828.

\bibitem{lomurno2021pareto}
E.~Lomurno, S.~Samele, M.~Matteucci, and D.~Ardagna, ``Pareto-optimal
  progressive neural architecture search,'' in \emph{Proceedings of the Genetic
  and Evolutionary Computation Conference Companion}, 2021, pp. 1726--1734.

\bibitem{falanti2022popnasv2}
A.~Falanti, E.~Lomurno, S.~Samele, D.~Ardagna, and M.~Matteucci, ``Popnasv2: An
  efficient multi-objective neural architecture search technique,'' in
  \emph{2022 International Joint Conference on Neural Networks (IJCNN)}.\hskip
  1em plus 0.5em minus 0.4em\relax IEEE, 2022, pp. 1--8.

\bibitem{falanti2022popnasv3}
A.~Falanti, E.~Lomurno, D.~Ardagna, and M.~Matteucci, ``Popnasv3: a
  pareto-optimal neural architecture search solution for image and time series
  classification,'' \emph{arXiv preprint arXiv:2212.06735}, 2022.

\bibitem{wu2019fbnet}
B.~Wu, X.~Dai, P.~Zhang, Y.~Wang, F.~Sun, Y.~Wu, Y.~Tian, P.~Vajda, Y.~Jia, and
  K.~Keutzer, ``Fbnet: Hardware-aware efficient convnet design via
  differentiable neural architecture search,'' in \emph{Proceedings of the
  IEEE/CVF Conference on Computer Vision and Pattern Recognition}, 2019, pp.
  10\,734--10\,742.

\bibitem{sandler2018mobilenetv2}
M.~Sandler, A.~Howard, M.~Zhu, A.~Zhmoginov, and L.-C. Chen, ``Mobilenetv2:
  Inverted residuals and linear bottlenecks,'' in \emph{Proceedings of the IEEE
  conference on computer vision and pattern recognition}, 2018, pp. 4510--4520.

\bibitem{howard2019searching}
A.~Howard, M.~Sandler, G.~Chu, L.-C. Chen, B.~Chen, M.~Tan, W.~Wang, Y.~Zhu,
  R.~Pang, V.~Vasudevan \emph{et~al.}, ``Searching for mobilenetv3,'' in
  \emph{Proceedings of the IEEE/CVF international conference on computer
  vision}, 2019, pp. 1314--1324.

\bibitem{hu2018squeeze}
J.~Hu, L.~Shen, and G.~Sun, ``Squeeze-and-excitation networks,'' in
  \emph{Proceedings of the IEEE conference on computer vision and pattern
  recognition}, 2018, pp. 7132--7141.

\bibitem{benyahia2019overcoming}
Y.~Benyahia, K.~Yu, K.~B. Smires, M.~Jaggi, A.~C. Davison, M.~Salzmann, and
  C.~Musat, ``Overcoming multi-model forgetting,'' in \emph{International
  Conference on Machine Learning}.\hskip 1em plus 0.5em minus 0.4em\relax PMLR,
  2019, pp. 594--603.

\bibitem{sarti2023anticipate}
\BIBentryALTinterwordspacing
S.~Sarti, E.~Lomurno, and M.~Matteucci, ``Anticipate, ensemble and prune:
  Improving convolutional neural networks via aggregated early exits,'' 2023.
  [Online]. Available: \url{https://arxiv.org/abs/2301.12168}
\BIBentrySTDinterwordspacing

\bibitem{howard2017mobilenets}
A.~G. Howard, M.~Zhu, B.~Chen, D.~Kalenichenko, W.~Wang, T.~Weyand,
  M.~Andreetto, and H.~Adam, ``Mobilenets: Efficient convolutional neural
  networks for mobile vision applications,'' \emph{arXiv preprint
  arXiv:1704.04861}, 2017.

\bibitem{deng2009imagenet}
J.~Deng, W.~Dong, R.~Socher, L.-J. Li, K.~Li, and L.~Fei-Fei, ``Imagenet: A
  large-scale hierarchical image database,'' in \emph{2009 IEEE conference on
  computer vision and pattern recognition}.\hskip 1em plus 0.5em minus
  0.4em\relax Ieee, 2009, pp. 248--255.

\bibitem{le2015tiny}
Y.~Le and X.~Yang, ``Tiny imagenet visual recognition challenge,'' \emph{CS
  231N}, vol.~7, no.~7, p.~3, 2015.

\end{thebibliography}

\end{document}